\documentclass[conference]{IEEEtran}

\IEEEoverridecommandlockouts

\usepackage{multirow}
\usepackage{multicol}
\usepackage[hyphens]{url}  
\usepackage{breakurl}

\usepackage{cite}
\usepackage{amsmath,amssymb,amsfonts}
\usepackage{algorithmic}
\usepackage{graphicx}
\usepackage{booktabs}
\usepackage{comment}
\usepackage{xcolor}
\usepackage{multirow}
\usepackage{float}
\usepackage{hyperref}
\usepackage{tikz}
\usepackage{csquotes}
\usepackage{xcolor}
\usepackage{lipsum}
\usepackage{subfigure}

\hypersetup{
    colorlinks=false, 
    pdfborder={0 0 1},
    linkbordercolor={0 1 0}, 
}

\begin{document}

\title{Benchmarking Large Language Models for Image Classification of Marine Mammals}

\author{Yijiashun Qi\textsuperscript{1}, Shuzhang Cai\textsuperscript{2}, Zunduo Zhao\textsuperscript{3}, Jiaming Li\textsuperscript{4}, Yanbin Lin\textsuperscript{5}, Zhiqiang Wang\textsuperscript{5+} \\
\textsuperscript{1}{\small University of Michigan}
\textsuperscript{2}{\small University of Texas at Dallas}
\textsuperscript{3}{\small New York University} \\
\textsuperscript{4}{\small Stony Brook University}
\textsuperscript{5}{\small Florida Atlantic University} \\
% \textsuperscript{*}{\small Equal Contribution}
\textsuperscript{+}{\small Corresponding Author}
% {\small Github \url{https://github.com/yeyimilk/LLMMarineMammal}}
}

\maketitle
\thispagestyle{plain}
\pagestyle{plain}

\begin{abstract}
    As Artificial Intelligence (AI) has developed rapidly over the past few decades, the new generation of AI, Large Language Models (LLMs) trained on massive datasets, has achieved ground-breaking performance in many applications. Further progress has been made in multimodal LLMs, with many datasets created to evaluate LLMs with vision abilities. However, none of those datasets focuses solely on marine mammals, which are indispensable for ecological equilibrium. In this work, we build a benchmark dataset with 1,423 images of 65 kinds of marine mammals, where each animal is uniquely classified into different levels of class, ranging from species-level to medium-level to group-level. Moreover, we evaluate several approaches for classifying these marine mammals: (1) machine learning (ML) algorithms using embeddings provided by neural networks, (2) influential pre-trained neural networks, (3) zero-shot models: CLIP and LLMs, and (4) a novel LLM-based multi-agent system (MAS). The results demonstrate the strengths of traditional models and LLMs in different aspects, and the MAS can further improve the classification performance. The dataset is available on GitHub: https://github.com/yeyimilk/LLM-Vision-Marine-Animals.git.
\end{abstract}


\begin{figure*}[th]
    \centering
    \subfigure{\includegraphics[width=0.32\columnwidth]{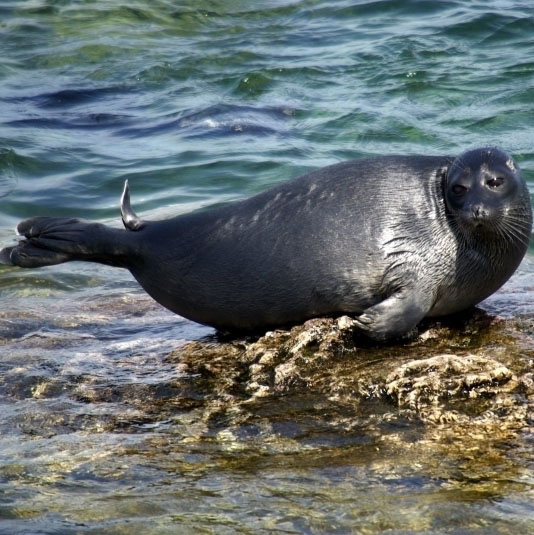}}
    \subfigure{\includegraphics[width=0.32\columnwidth]{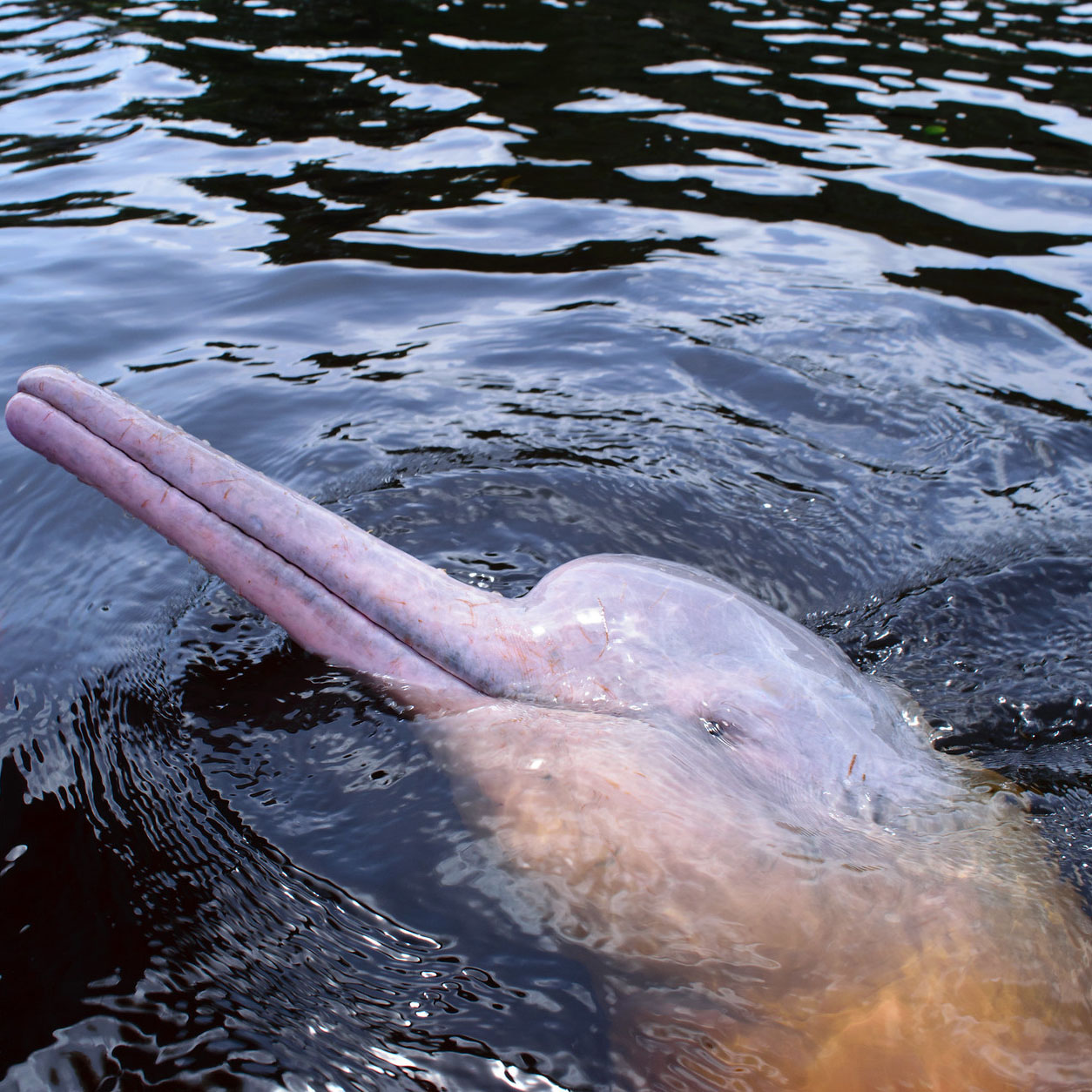}}
    \subfigure{\includegraphics[width=0.32\columnwidth]{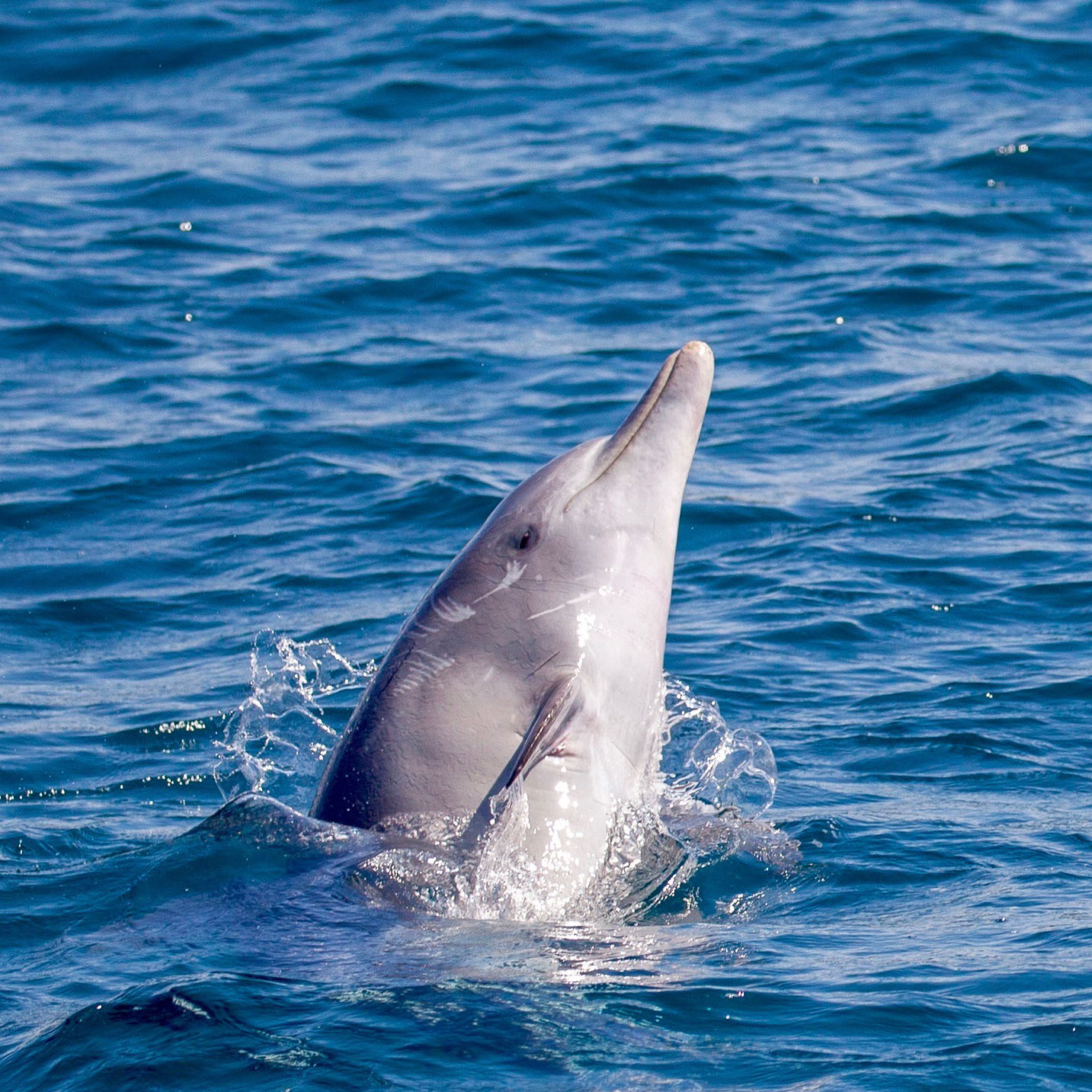}}
    \subfigure{\includegraphics[width=0.32\columnwidth]{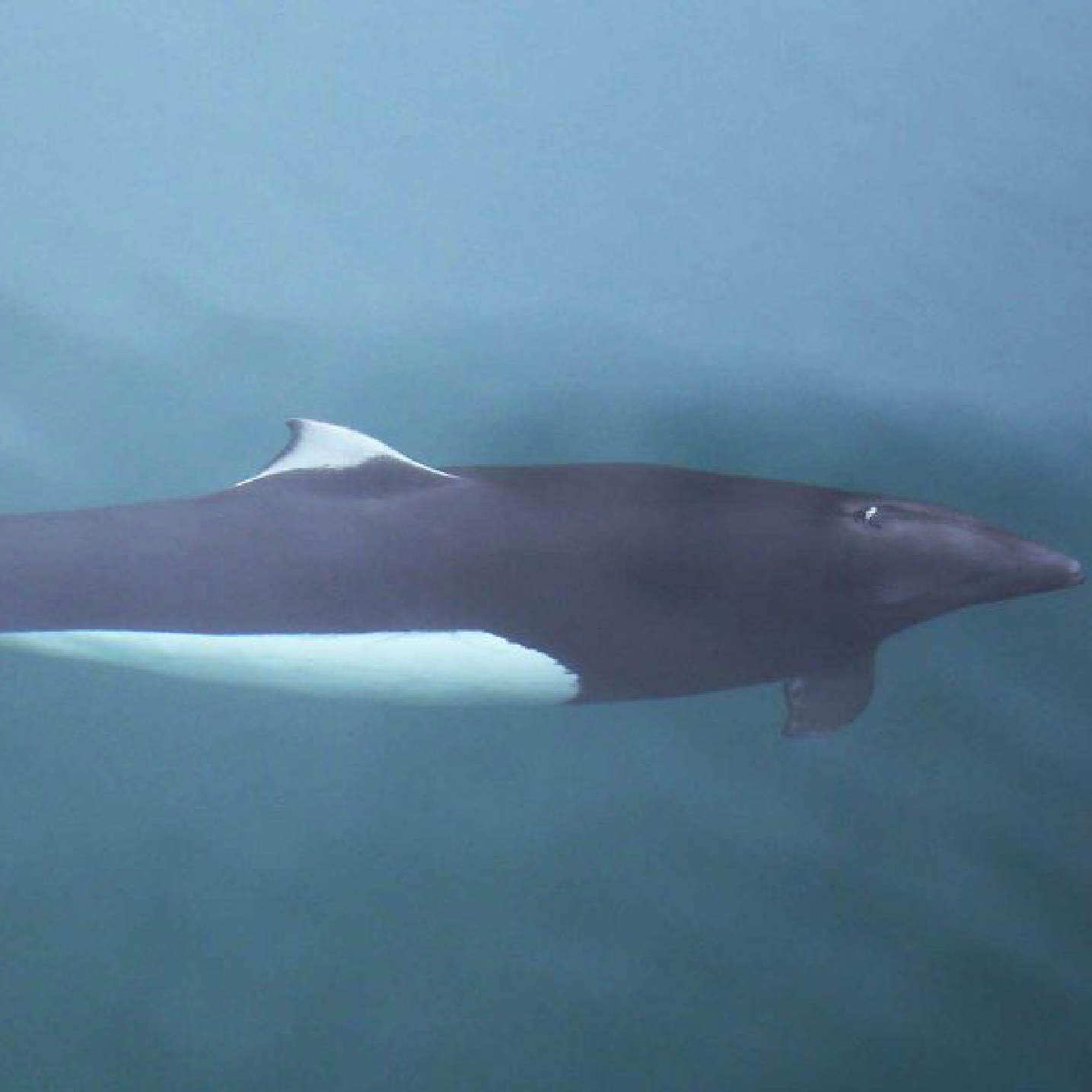}}
    \subfigure{\includegraphics[width=0.32\columnwidth]{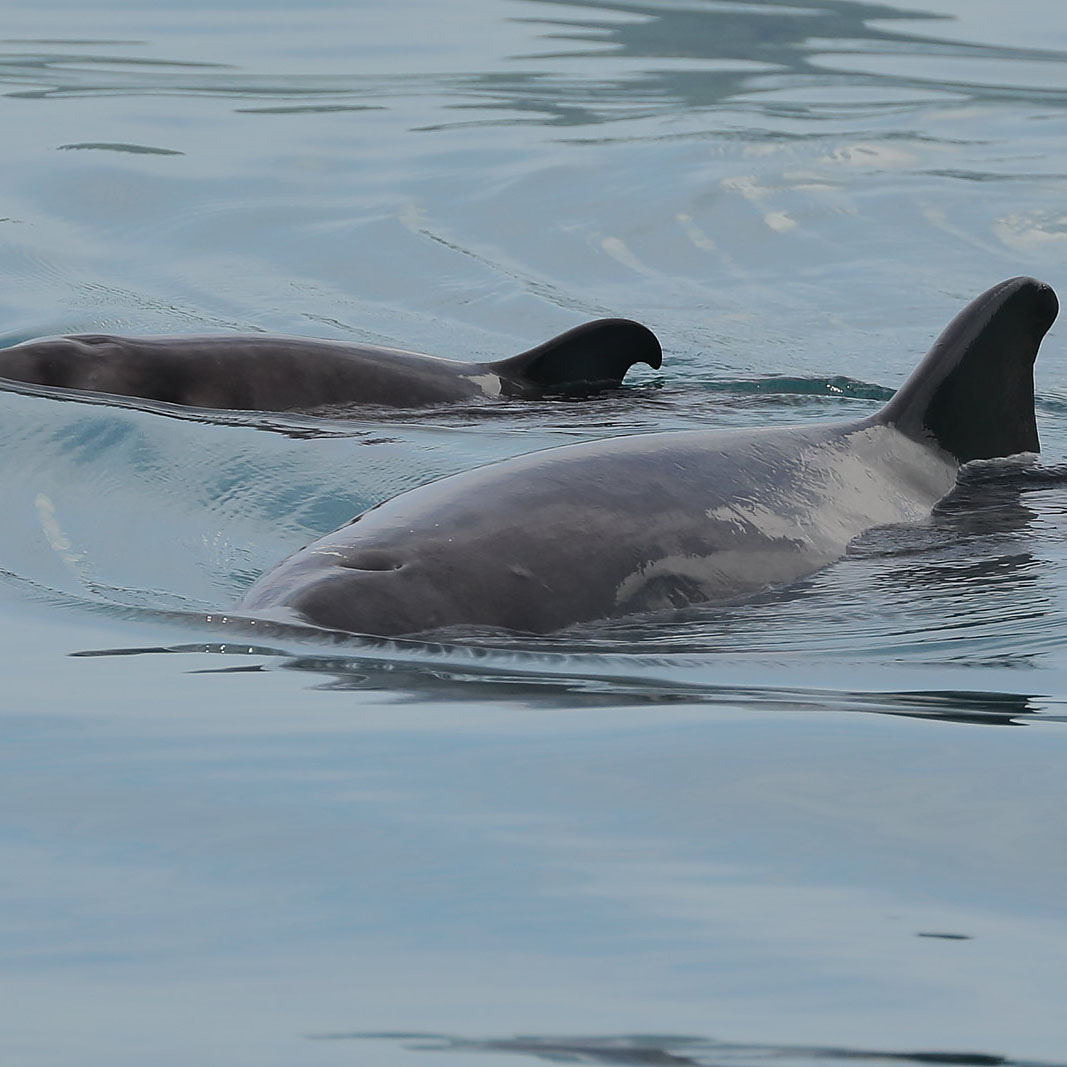}}
    \subfigure{\includegraphics[width=0.32\columnwidth]{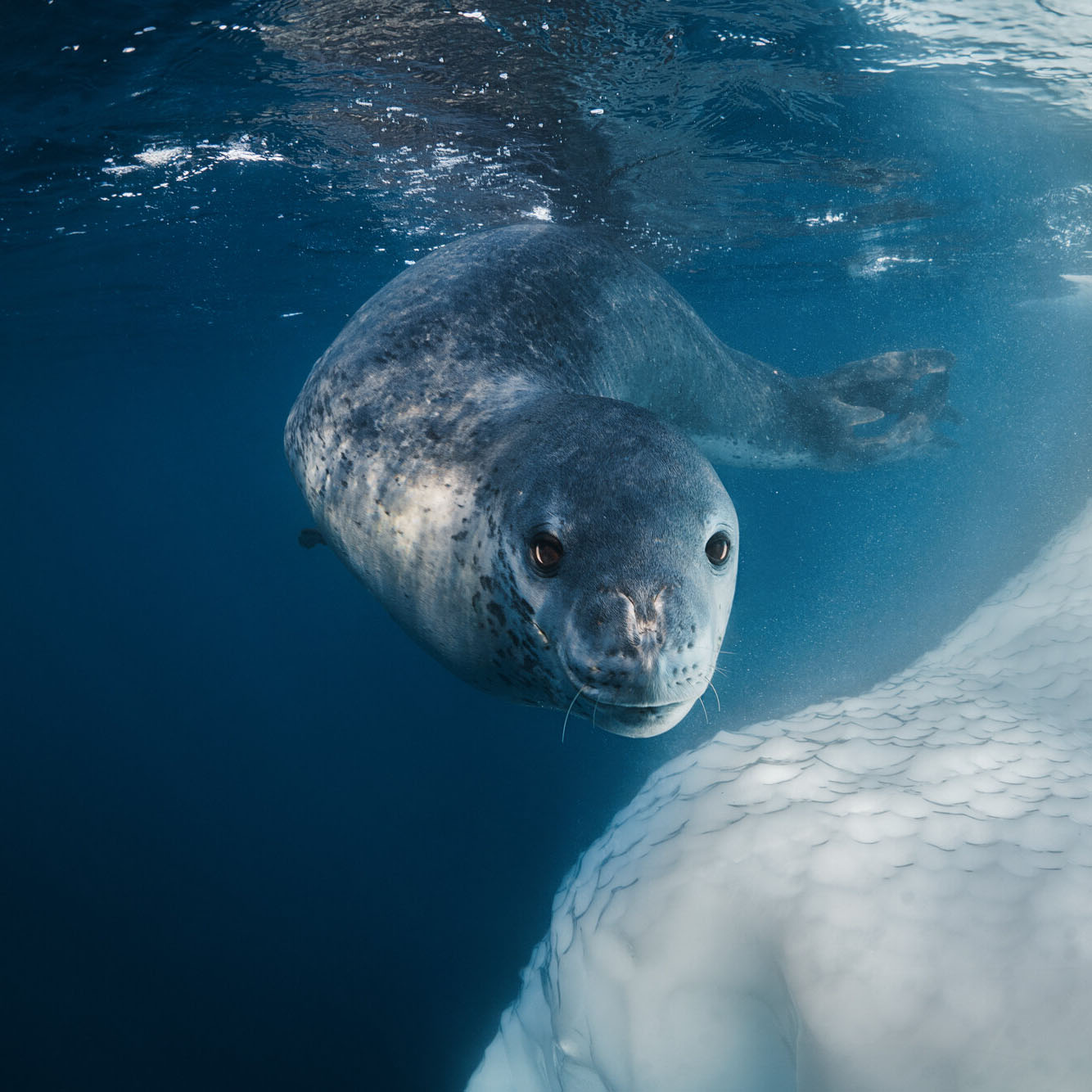}}
    \subfigure{\includegraphics[width=0.32\columnwidth]{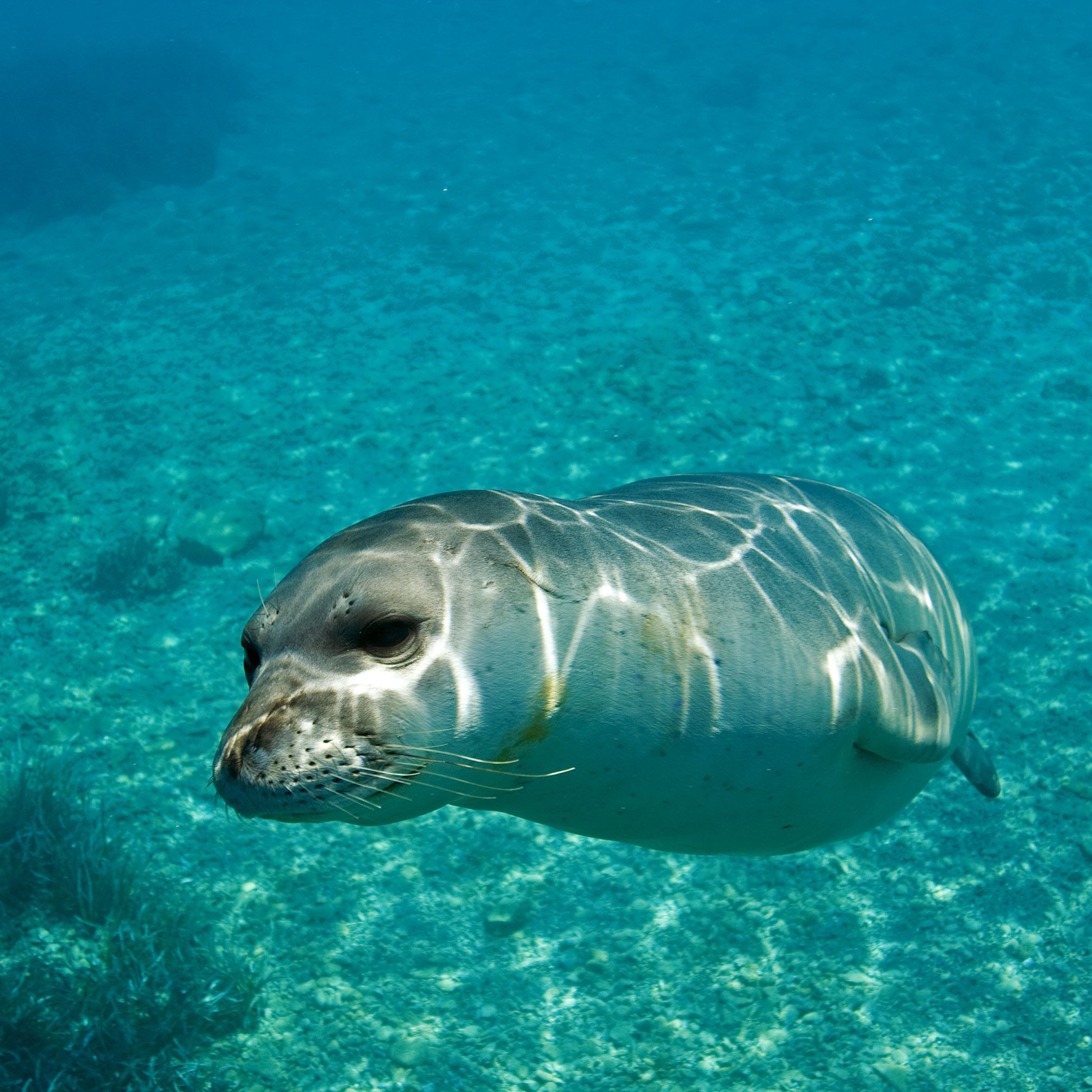}}
    \subfigure{\includegraphics[width=0.32\columnwidth]{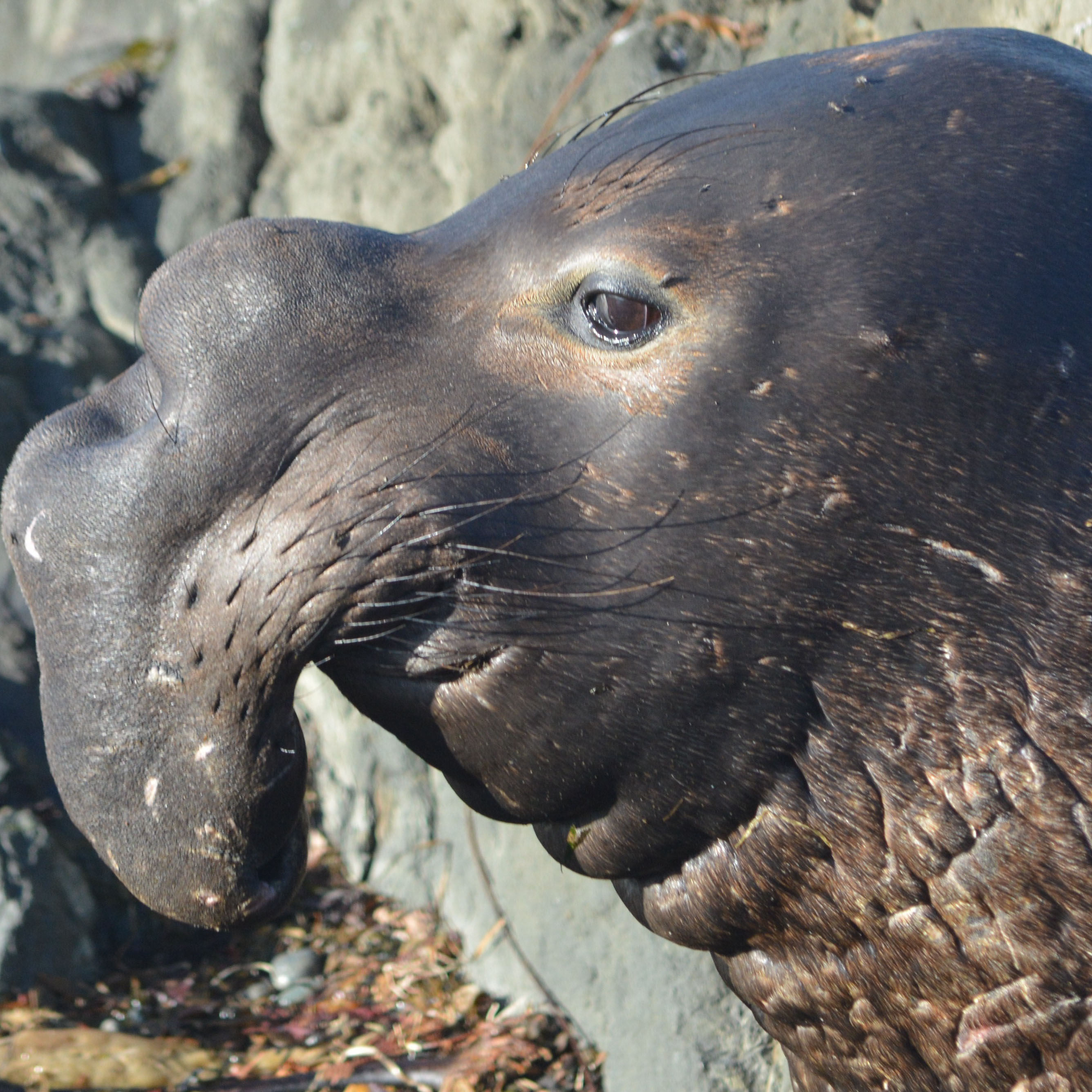}}
    \subfigure{\includegraphics[width=0.32\columnwidth]{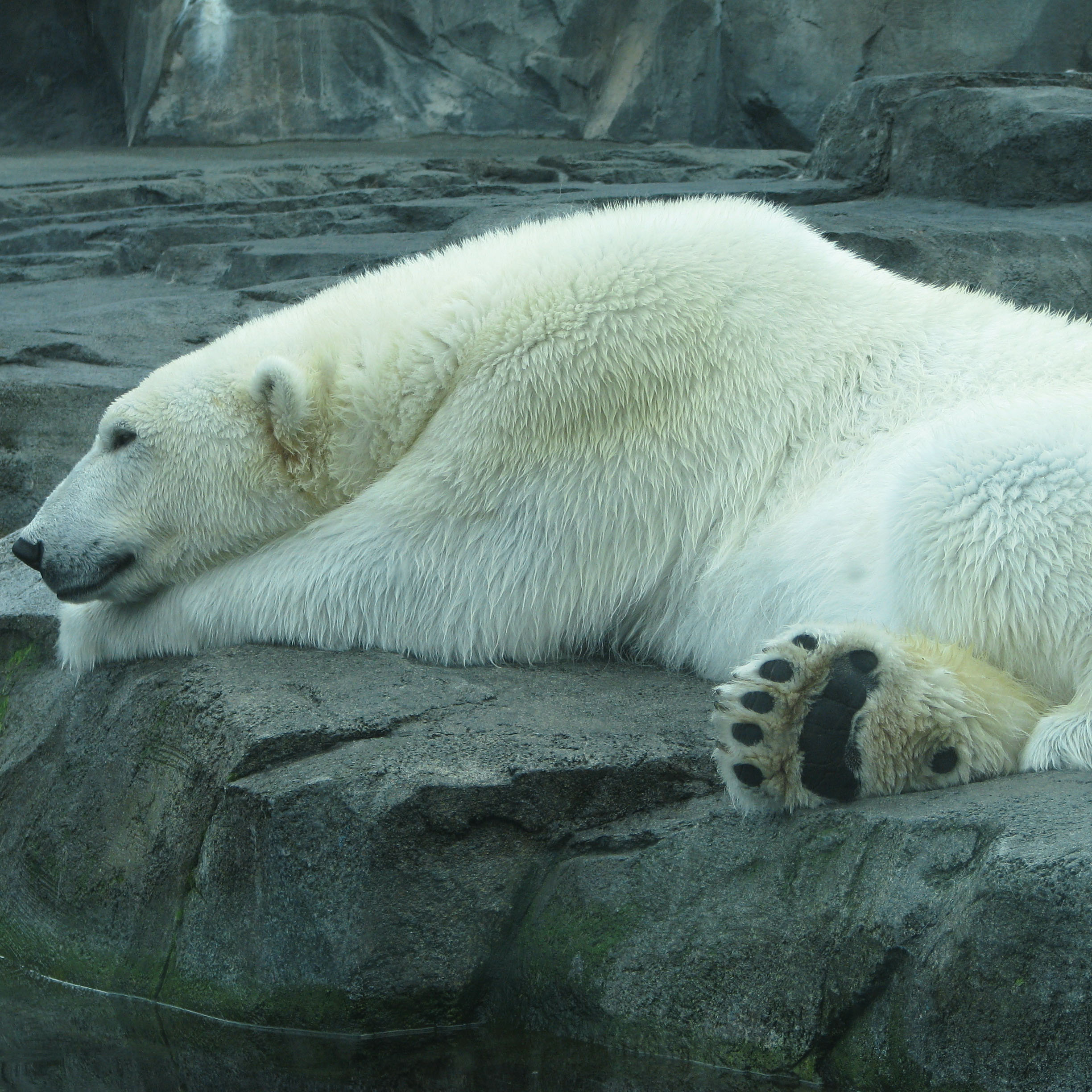}}
    \subfigure{\includegraphics[width=0.32\columnwidth]{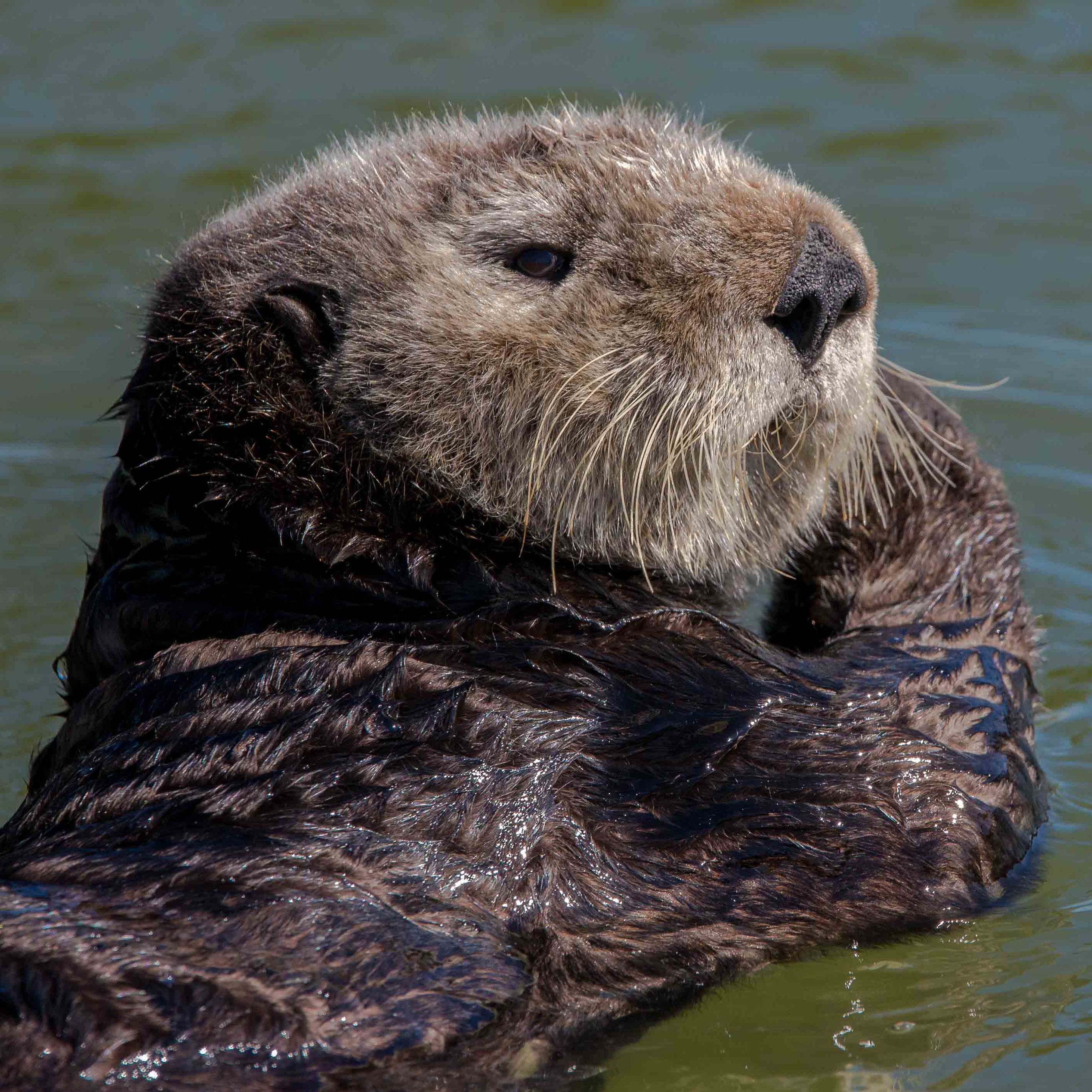}}
    \subfigure{\includegraphics[width=0.32\columnwidth]{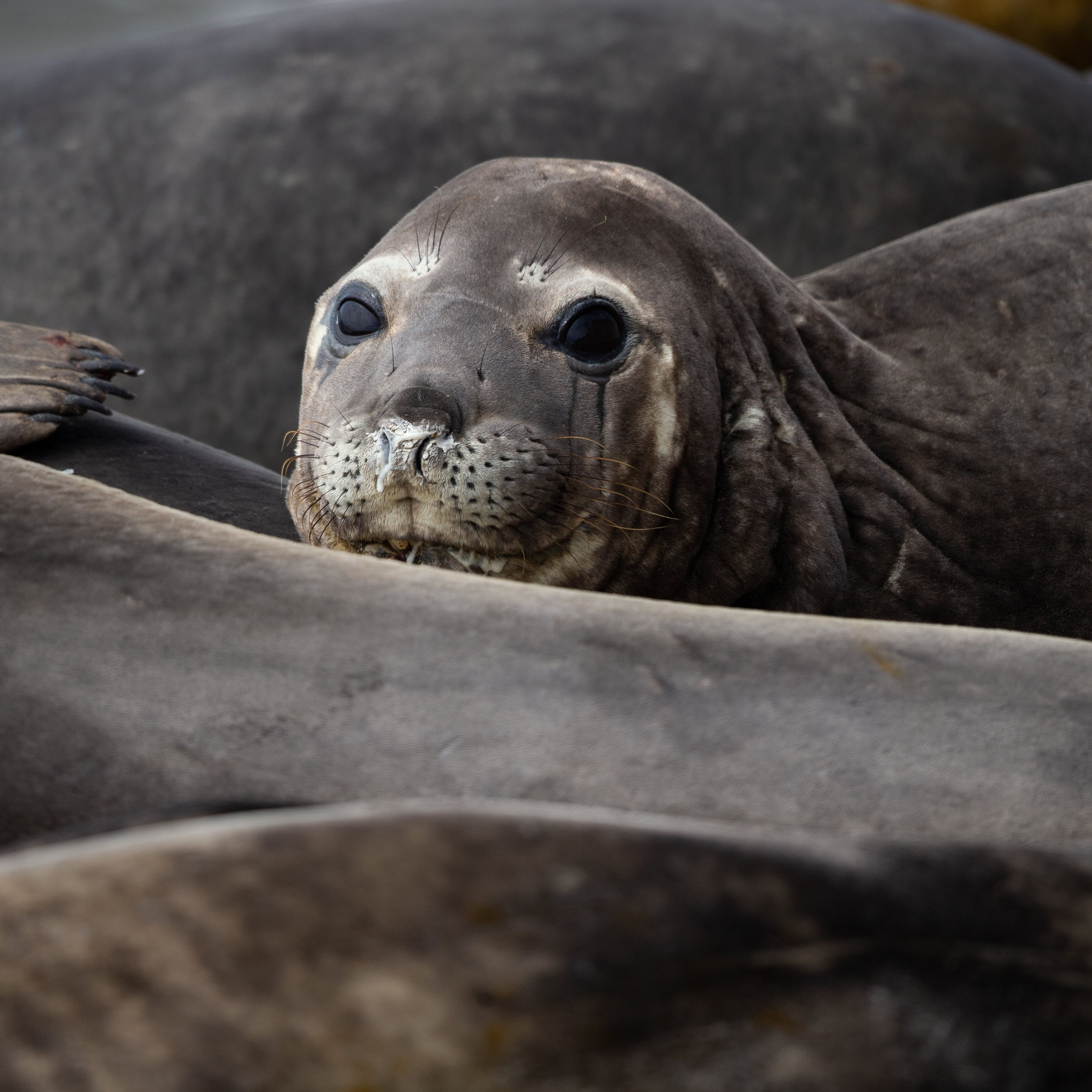}}
    \subfigure{\includegraphics[width=0.32\columnwidth]{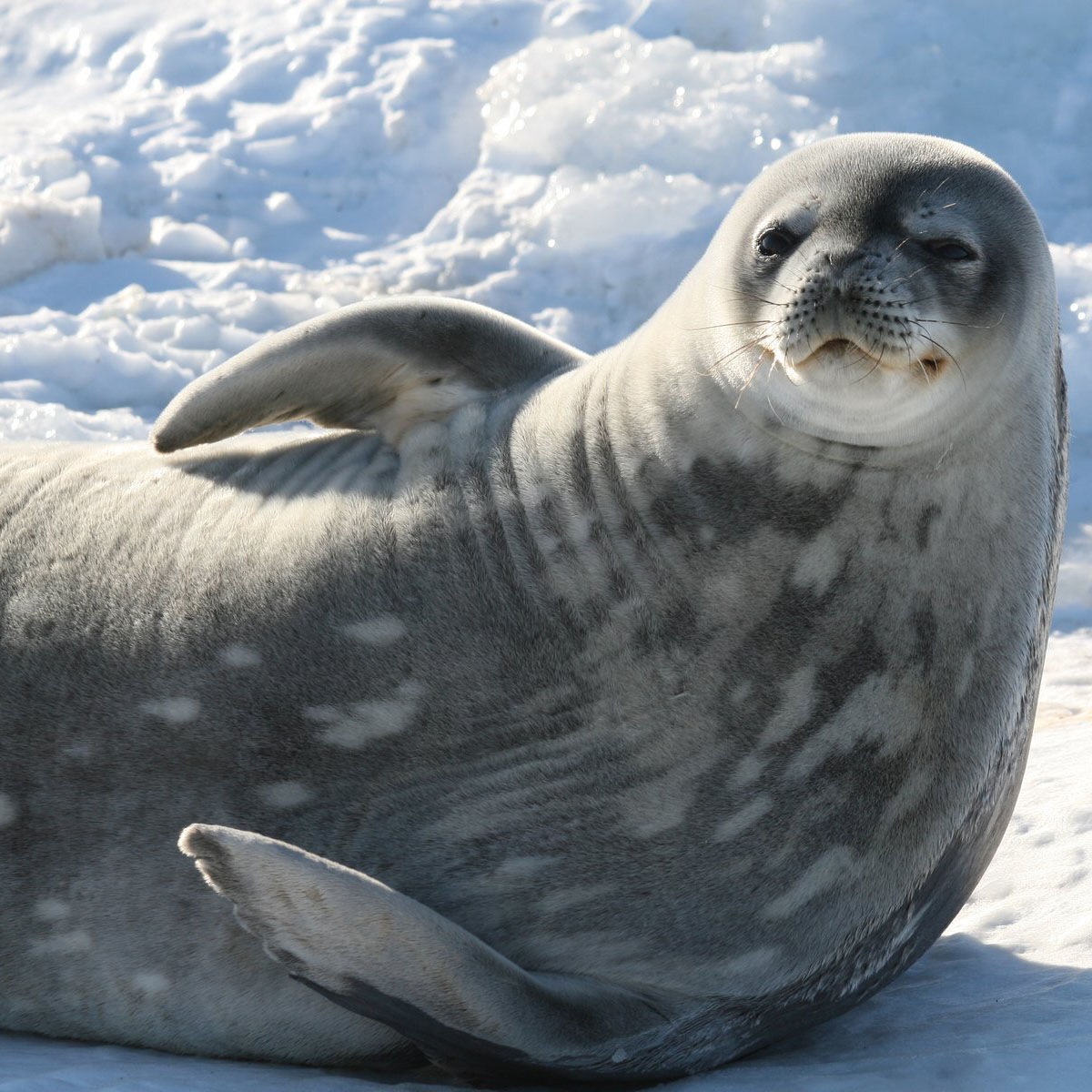}}
    \caption{Image samples from the proposed dataset.}
    \label{fig: sample}
\end{figure*}

\section{Introduction}
Marine mammals are integral to ocean environments and are critical for maintaining ecological balance \cite{bowen1997role, moore2008marine}. However, they are increasingly threatened by habitat destruction, climate change, pollution, and various human activities. These challenges underscore the urgent need for extensive research and targeted conservation efforts \cite{nelms2021marine}. A growing trend is to utilize artificial intelligence (AI) to study nature more effectively; in this paper, we investigate and enhance the capabilities of the most influential models for marine mammal classification.
% Artificial intelligence has shown remarkable advancements in recent years, particularly in developing large language models (LLMs). These models, such as GPT-4, Llama3, and Claude, have demonstrated exceptional capabilities in understanding and handling various text tasks, such as text classification, summarization, question answering, code generation, etc. By integrating vision and language in multimodal frameworks, LLMs with vision components have been shown to understand and interpret images with textual information, enabling a more nuanced and comprehensive visual content analysis, for example, image-based geolocation, image captioning, and vision question answering. 

The research into AI has made remarkable progress in recent years, particularly in the development of large language models (LLMs) such as GPT \cite{achiam2023gpt, gpt4o2024}, Gemini \cite{team2023gemini, reid2024gemini} and LLaMA \cite{touvron2023llama, touvron2023llama2}. These models have demonstrated exceptional proficiency across a wide range of text-related tasks, including text classification, summarization, question answering, and code generation. Recent advancements have further expanded the capabilities of these models by incorporating vision recognition into multimodal frameworks, making them able to analyze and interpret images alongside textual data. There are two types of multimodal LLMs: the first includes those specifically designed for visual tasks, such as LLaVA \cite{LLaVA}, DALL-E 3 \cite{DALLE3}, BLIP-2 \cite{li2023blip}, CLIP \cite{CLIP}, and ALIGN \cite{ALIGN}. The second kind includes general LLMs, such as GPT-4o \cite{gpt4o2024} and Gemini-1.5 Pro \cite{gemini1.5pro2024}, which integrate visual processing functions into their text generation capabilities. This integration facilitates more sophisticated and comprehensive analyses of visual content, enabling a series of image-related tasks \cite{singh2024evaluating, wang2024llmgeo}. These capabilities provide a richer, more context-aware approach to processing multimodal data.

Despite the rapid advancements in LLMs, the availability of specialized datasets for training and evaluating these models remains inadequate. While existing datasets encompass a broad range of categories, from generic objects to specific groups such as animals, there is a notable deficiency in datasets explicitly focused on marine mammals. Furthermore, there is a significant lack of benchmarks designed to assess the performance of LLMs on such specialized data as marine mammals. This scarcity presents a substantial barrier to conducting detailed analyses and developing conservation strategies tailored to these critical species.

\begin{figure*}[!t]
    \centering
    \includegraphics[width=0.80\textwidth]
     {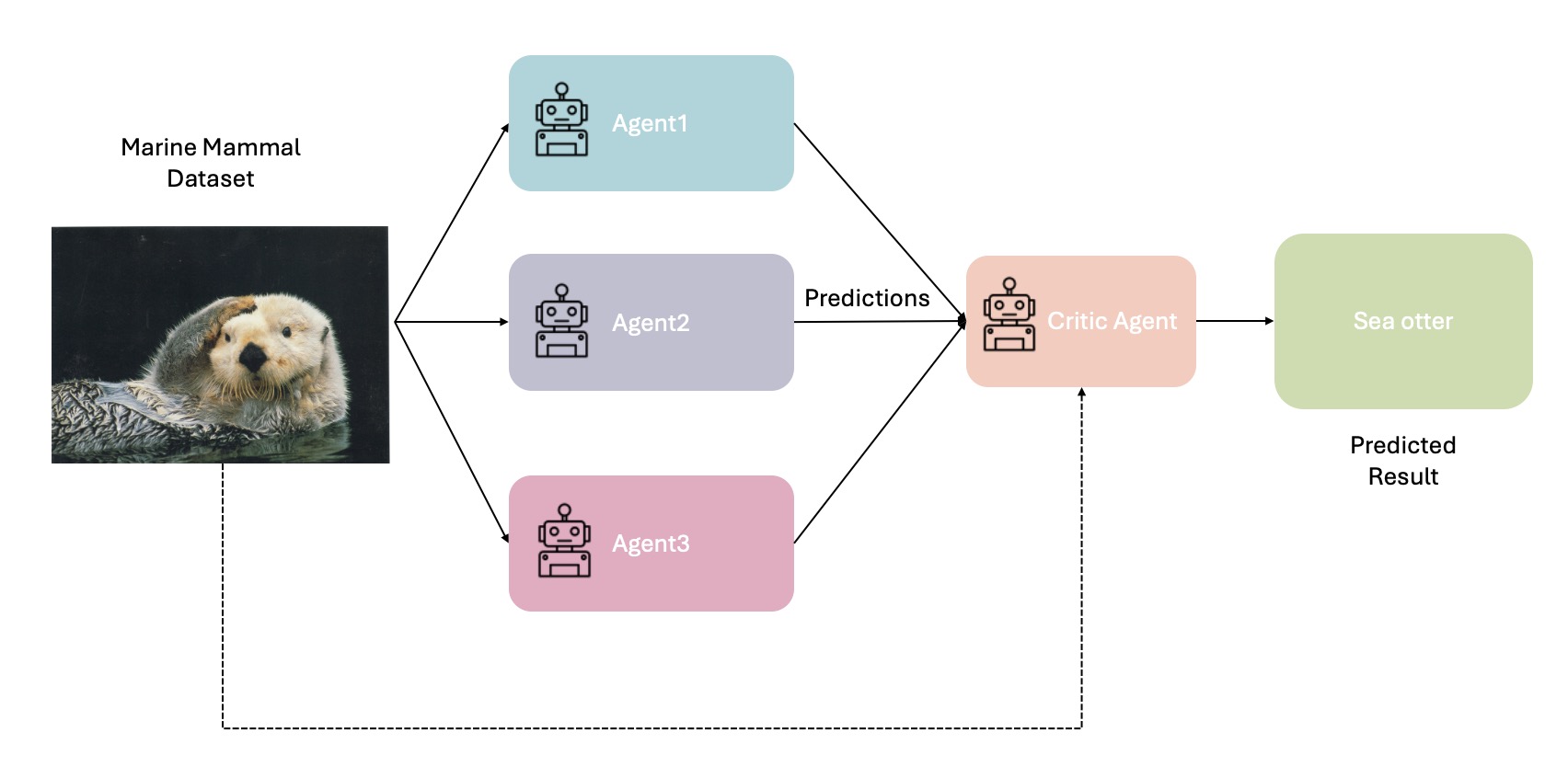}
    \caption{The process of the LLM-based MAS.}
    \label{fig:multi-agent-system}
\end{figure*}

To address these limitations, our work introduces the following major contributions:

\begin{itemize}
    \item \textbf{A Novel Dataset:} We introduce a dataset specifically focused on marine mammals, comprising a diverse array of images accompanied by corresponding labels at various levels of granularity. This dataset is designed to represent a wide range of species, capturing them across different environments, behaviors, and group dynamics.
    \item \textbf{Comprehensive Model Evaluations:} We conduct thorough evaluations of model performance, including traditional machine learning (ML) techniques, deep learning (DL) approaches, zero-shot learning models such as Contrastive Language-Image Pretraining (CLIP) and multimodal LLMs. Our evaluations explore various strategies and dimensions to rigorously assess the capabilities and limitations of these models. Through these contributions, we aim to address existing gaps and advance research in marine mammal studies while also contributing to the development of the latest LLMs.
    \item \textbf{An innovative multi-agent system:} We propose and develop an LLM-based multi-agent system (MAS) to further explore LLMs' potential in marine mammal classification. The system takes each foundational LLM's prediction as input for an ensemble LLM, which incorporates diverse reasoning chains and provides a more comprehensive analysis. The outcomes indicate that our MAS outperforms a single LLM at different levels of classification. The proposed multi-agent system is demonstrated in Figure \ref{fig:multi-agent-system}.
\end{itemize}

\section{Related Work}
\subsection{Research on Marine animals in Aquatic Environments}
Marine mammals, encompassing species such as whales, dolphins, seals, and sea lions, play a crucial role in ocean ecosystems \cite{bowen1997role,moore2008marine}. However, the conservation of these species is challenged by the complexity of their natural habitats and the threats they face. The application of Computer Vision (CV) to marine research is a growing area of interest \cite{pedersen2019detection,wang2023deep}, yet it remains underdeveloped, particularly concerning datasets focused exclusively on marine mammals. Previous research in aquatic environments has predominantly centered on fish detection, driven by the needs of the aquaculture industry. However, the detection and analysis of marine mammals remain an area requiring further exploration.

Moreover, local regression models applied to aquatic species face numerous challenges, including varying illumination, low contrast, high noise levels, species deformation, frequent occlusion, and dynamic backgrounds \cite{yang2020cv_aquaculture, zhang2023sensors}. These challenges are exacerbated in marine environments, where conditions can be highly variable and difficult to control.

\subsection{AI Techniques in Animal Recognition and Data Scarcity for Marine Mammals}

AI techniques, particularly those involving deep learning, have been successfully applied to various animal recognition tasks. In the domain of terrestrial animals, approaches such as Convolutional Neural Networks (CNNs) \cite{ibraheam2021performance} have demonstrated high accuracy and efficiency. For example, a study on cattle brand recognition \cite{silva2017cattle} used CNNs for feature extraction and SVMs for classification, achieving an accuracy of $93.28\%$ and indicating the potential of these methods for similar applications. 

Recent advancements in LLMs, particularly in integrating vision and language into multimodal models, offer promising new directions for marine mammal research. These models, which process both textual and visual data, have the potential to overcome some of the challenges posed by traditional CV techniques. For instance, models like Gemini-1.5 Pro have shown proficiency in zero-shot learning tasks, where the model can recognize and categorize images without prior exposure to specific labels \cite{reid2024gemini}.

Despite these advancements, the application of advanced techniques to marine mammal tasks remains underexplored. The lack of benchmarks and specialized datasets for evaluating these models on marine mammal data has hindered progress in this area\cite{jian2024underwater}. To address this gap, we introduce a new dataset focused exclusively on marine mammals. Additionally, our comprehensive evaluation of model performance across traditional ML models, pre-trained DL models, LLMs and an LLM-based MAS provides valuable insights into the capabilities and limitations of these advanced AI techniques in the context of marine mammal research.

\subsection{LLMs-based Multi-Agent Systems}

LLMs have shown remarkable abilities in understanding and processing complex reasoning tasks. Recent studies suggest that these strengths can be significantly enhanced when deployed within a Multi-Agent System in different applications \cite{hong2023metagpt, wang2023avalon, park2023generative, zeng2024autodefense}. In an LLMs-based MAS, multiple LLMs function as independent agents, each assigned a specialized role. These agents work in concert to address tasks that may be too complex for a single model to manage effectively. The system capitalizes on the collaborative and complementary strengths of these agents, enabling them to engage in planning, discussion, and decision-making processes that closely resemble those of humans \cite{sreedhar2024simulating} or teams \cite{guo2024embodied}.

The key strengths of LLMs-based MAS can be in several aspects. Its ability to specialize tasks allows each agent to focus on a specific aspect of the problem and leads to more precise and effective solutions. The diversity of thought provided by multiple agents ensures that various perspectives are considered, enhancing the overall problem-solving process. Additionally, the system’s collaborative nature promotes robust error detection and correction, as agents can cross-check and validate each other’s outputs. LLMs-based MAS is increasingly becoming a powerful tool for tackling complex, multifaceted challenges \cite{guo2024large}.

\section{Dataset}

\begin{figure*}[h]
    \centering
    \includegraphics[width=1.0\textwidth]
    {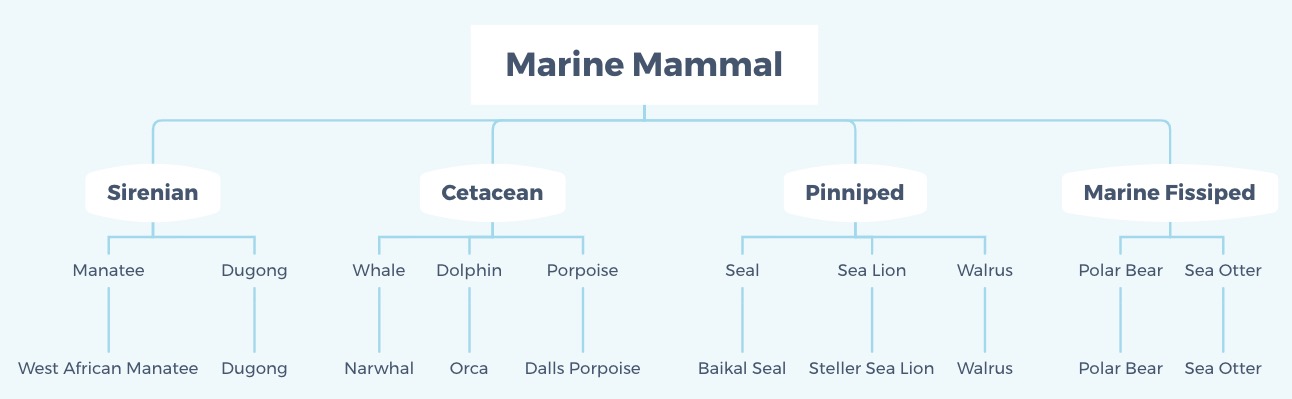}
    \caption{The figure shows the hierarchical structure of the labels used in the marine mammal dataset. There are three levels: 4 groups, 10 medium sub-groups, and 65 species, with each medium sub-group represented by 1 species. }
    \label{fig:label_hierarchy}
\end{figure*}

As part of preparing a comprehensive framework for image classification, this paper introduces a meticulously curated dataset of marine mammal images sourced from various platforms. These include \emph{Flickr}\footnote{\url{https://www.flickr.com/}}, \emph{Getty Images}\footnote{\url{https://www.gettyimages.com/}}, \emph{GBIF}\footnote{\url{https://www.gbif.org/}}, \emph{World Register of Marine Species (WoRMS)}\footnote{\url{https://www.marinespecies.org/aphia.php}}, \emph{Unsplash}\footnote{\url{https://unsplash.com/}}, and \emph{“OCEAN TREASURES” Memorial Library}\footnote{\url{https://otlibrary.com/}}. By using species names as the primary search keywords on these platforms, the resulting dataset is both diverse and inclusive in the representation of marine mammals.

\begin{table}[h]
 \caption{Statistics of the dataset}
 \renewcommand{\arraystretch}{1.2}
  \tabcolsep 9pt
 \label{tab:dataset_stat2}
 \centering
 \begin{tabular}{l l l l} 
 \toprule
  & Training set & Validation set & Testing set \\ 
 \midrule 
  \# images & 1163 & 130 & 130 \\
  Min \# per species & 11 & 2 & 2 \\
  Max \# per species & 37 & 2 & 2 \\
  \# Group classes & 4 & 4 & 4 \\
  \# Medium classes & 10 & 10 & 10 \\
  \# Species classes & 65 & 65 & 65 \\
 \bottomrule
 \end{tabular}
\end{table}

Overall, the dataset consists of 1,423 images of 65 kinds of marine mammals, providing a comprehensive view of marine biodiversity. Figure \ref{fig: sample} displays random sample images from the proposed dataset. The dataset includes a variety of viewpoints, such as underwater, at the water's surface, aerial views, from a boat, inside aquariums, and along the shorelines. The animals in the image may be fully submerged, partially in and out of the water, or completely exposed. Air and water conditions vary depending on the time and location where the images were captured. The images remain unprocessed, preserving their original quality to exhibit real-world use cases without pre-selection and pre-processing.

Table \ref{tab:dataset_stat2} provides a statistical overview of the dataset, elucidating the distribution of images among the three subsets: training, validation, and testing. All subsets include all classes for different levels. Each species is represented by 2 images in the validation and testing set, and for the training set, the number ranges from 11 to 37, with most species having 16 images. Each image is carefully evaluated for its relevance and quality, ensuring the rigorous nature of the dataset for future model development and comparative studies.

In this dataset, each image is classified into three hierarchical levels: 4 group-level classes, 10 medium-level classes, and 65 species-level classes. Specifically, every image has a distinct label corresponding to each classification level. The label hierarchy is illustrated in Figure \ref{fig:label_hierarchy}.

\section{Experimental Settings}

In this section, we introduce the settings used in our experiments. The classification of marine mammal images is a complex task due to the diverse range of species, their similar appearances, and the varying environmental conditions in which images are captured. To achieve a more comprehensive evaluation, we assess these models at different levels of granularity: species, medium, and group. The objectives of these experiments are to establish the baseline performance of traditional ML models and pre-trained ML models and to investigate the performance of advanced multimodal LLMs and a novel LLM-based multi-agent system for the task.

\subsection{Overview of Model Categories}

To conduct a thorough evaluation, we select models from three main categories: traditional ML models, pre-trained DL models, and multi-modal LLMs, as follows. 

\begin{itemize}
    \item \textbf{ML models, KNN and SVM}: In this scheme, a neural network is used as a feature extractor, followed by traditional machine learning classifiers such as K--Nearest Neighbors (KNN) and SVM. It serves as a baseline to reflect the performance of more complex models.
    \item \textbf{Pre-trained DL models, VGG\cite{simonyan2014very} and ResNet\cite{he2016deep}}: VGG is well-known for its excellent performance and is widely used as a benchmark standard for DL models. ResNets learn residual mappings instead of direct mappings from input to output, which enables the construction of very deep networks, with some versions containing hundreds or even thousands of layers.
    \item \textbf{CLIP\cite{CLIP}}: 
    CLIP (Contrastive Language–Image Pre-training) builds on a large body of work on zero-shot transfer, natural language supervision, and multimodal learning. It was trained on 400 million (image, text) pairs collected from the internet and can be applied to any visual classification benchmark by simply providing the names of the visual categories to be recognized, referred as zero-shot capabilities.
    \item \textbf{LLaVA}\cite{LLaVA2023visual}: The Large Language and Vision Assistant (LLaVA) represents a multimodal model combining a vision encoder and a language model, enabling both visual and language understanding. It has been effective in tasks requiring visual instruction, leveraging visual features to generate natural, human-like language responses.
    \item \textbf{GPT-4o}\cite{gpt4o2024}: A new variant of GPT-4 optimized for vision and audio tasks. GPT-4o also leverages large-scale pretraining datasets, improving performance in complex reasoning tasks and other advanced natural language processing challenges.
    \item \textbf{Gemini-1.5 Pro} \cite{gemini1.5pro2024}: As one of the latest models from Google DeepMind, Gemini-1.5 Pro is designed and trained to handle a broad range of tasks, providing state-of-the-art results in both natural language and multimodal tasks.
\end{itemize}

\subsection{Evaluation Metrics}

In this study, accuracy is chosen as the primary metric for assessing the performance of different models, and it is defined as shown in Equation \ref{equ: acc} below:

\begin{equation}
Accuracy = \frac{\sum_{1}^{n}TP_{i}}{N}
\label{equ: acc}
\end{equation}

where: $TP_{i}$ is the number of true positives for class $i$, $n$ is the number of classes, and $N$ is the total number of predictions.

Other metrics are not employed as they may not reflect the performance well. For example, the micro-F1 is always equal to accuracy in multi-class classification, as there are no true negatives; every sample belongs to exactly one class.

We use accuracy alone to provide a clear and direct evaluation of the model's ability to distinguish between different marine mammal species without overcomplicating the analysis. This choice aligns well with the dataset's balanced characteristics and the specific research objectives of this study.

\subsection{Machine Learning settings}
In order to generate a performance baseline using traditional ML methods, we employ the pre-trained model DINOv2 to generate embeddings for images in our dataset. Our decision to focus on DINOv2 is supported by recent research, which demonstrates that DINOv2, in its latest version, outperforms CLIP in generating robust and transferable visual features \cite{oquab2023dinov2}. The embeddings are extracted by passing the images through the pre-trained model. They capture the underlying patterns and characteristics and serve as the representations of images in the latent feature space. The extracted features are then used to train KNN and SVM classifiers on the training set. 

\subsection{Deep Learning settings}
In the deep learning experiments, we employ ResNet and VGG families to classify marine mammal images using a fine-tuning approach. Pre-trained versions of these models are obtained from the PyTorch library \cite{paszke2019pytorch}.

During the training process, we compare ResNet models of varying depths: ResNet-34, ResNet-50, ResNet-101, and ResNet-152. Similarly, in the VGG model family, we fine-tune VGG-11 and VGG-13, each with and without batch normalization. A new fully connected layer is added as the final layer to predict the likelihood of each class. We use pre-trained models from the PyTorch library and leverage the Adam optimizer to tune the network parameters over 40 epochs at the species, medium, and group levels, respectively. After each epoch, the fine-tuned models are evaluated on the validation set. Within each model family, models with the highest validation accuracy are selected and compared (discussed in \textit{Section V}). The best-performing model is then evaluated on the test set to determine the accuracy. If two models achieve the same validation accuracy, we take the smaller model.

\subsection{Multimodal LLMs}

We use four base models: LLaVA-v1.5-7B, LLaVA-v1.6-34B, GPT-4o, and Gemini-1.5-Pro, and evaluate them using the prompts stated in \hyperref[Prompting_Strategies]{\textit{Prompting Strategies}}. We use a zero-shot approach where the models classify marine mammal images directly without additional training. These models leverage both visual and textual information to make predictions, making them suitable for scenarios where labeled training data is sparse. To optimize their performance, we design a set of prompting strategies that guide the models through different levels of specificity and reasoning. The models are tested using eight unique prompts that combine different question formats and answer granularities.

\subsection{Prompting strategies}

\label{Prompting_Strategies}
The prompting strategies are crafted to optimize the LLMs' performance and are presented below. They are divided into two main parts: the question format (Part 1) and the answer granularity (Part 2). The question format includes two types: Zero-Shot (0s), which asks the model to directly identify the species without any intermediate reasoning steps; and Zero-Shot Chain-of-Thought (0s-CoT), which guides the model to think step-by-step before reaching the final decision. The answer granularity focuses on different levels of classification --- species-level, medium-level, and group-level --- using either open-ended prompts or predefined lists of possible answers. By combining these parts, we create eight unique prompts for the LLMs.

\noindent
\textbf{a): Part 1: Question Format}
\begin{itemize}
    \item \textbf{Zero-Shot (0s)}: 
    \textit{"What is the species of the marine mammal(s) in the picture? Provide only the direct answer."}
    This prompt asks the model to directly identify the species without any intermediate reasoning steps. It is designed to test the model's ability to provide an immediate response based on its pre-trained knowledge.
    \item \textbf{Zero-Shot Chain-of-Thought (0s-CoT):} 
    \textit{"What is the species of the marine mammal(s) in the picture? Let's think step by step."}
    This prompt asks the model to break down the problem and provide reasoning steps using the CoT method. It aims to leverage the model's reasoning capabilities.
\end{itemize}

\noindent
\textbf{b): Part 2: Answer Granularity}
\begin{itemize}
    \item \textbf{Open-Ended Species-Level}: 
    \textit{"The answer must be the common species name, avoiding any scientific or higher-level classifications."}
    This prompt instructs the model to provide the open-ended answer in species-level classification.
    \item \textbf{Name-Listed Species-Level}: 
    \textit{"The answer must be one from the following: [List of species names]."}
    This prompt provides a predefined list of species names, guiding the model to choose from the list at the species level.
    \item \textbf{Name-Listed Medium-Level}: 
    \textit{"The answer must be one from the following ten taxonomic groups: Dolphin, Whale, Porpoise, Seal, Sea Lion, Walrus, Manatee, Dugong, Polar Bear, and Sea Otter."}
    This prompt asks the model to classify the marine animal into one of ten medium-level taxonomic groups. 
    \item \textbf{Name-Listed Group-Level}: 
    \textit{"The answer must be one from the following four taxonomic groups: Cetacean, Pinniped, Sirenian, and Marine Fissiped."}
    This prompt asks the model to classify the marine animal into one of four broad taxonomic groups.
\end{itemize}

\subsection{LLM-based MAS}
To further improve the performance of LLMs, we construct a MAS that leverages the outputs from three foundational LLMs --- GPT-4o, Gemini-1.5 Pro, and LLaVA-34B. In this setup, each LLM acts as an independent agent, providing a primary decision based on its own capabilities. These primary predictions are then aggregated into an ensemble model (referred to as the Critic Agent in Figure \ref{fig:multi-agent-system}) --- GPT-4o or Gemini-1.5 Pro --- to derive the final classification result.

The MAS experiment settings include two scenarios: with and without the use of CoT reasoning. Under the CoT setting, reasoning chains are incorporated into the primary predictions, allowing the ensemble model to consider more complex decision-making processes.
\begin{table}[ht]
    \centering
    \begin{small}
    \renewcommand{\arraystretch}{1.3}
    \tabcolsep 12pt
    \caption{\small{Accuracy Table}}
    \begin{tabular}{l l l l}
   % \begin{tabular}{p{2cm} p{0.5cm} p{0.8cm} p{0.8cm} p{0cm}}
    \hline
        \multirow{2}{*}{Method} & \multicolumn{1}{ c }{Species} & \multicolumn{1}{ c }{Medium} & \multicolumn{1}{ c }{Group} \\ 
        \cline{2-4} & ACC & ACC & ACC \\ 
        \hline

        KNN  & $43.85$ & $84.62$  & $98.46$  \\
        SVM  & $\mathbf{64.62}$ & $90.00$  & $\mathbf{100.00}$ \\ 
        
        \hline
        
        VGG-11 w/ BN & $60.77$ & $86.15$  & $93.85$  \\
        ResNet-50  & $60.00$ & $86.15$  & $98.46$  \\ \hline
        
        CLIP & $36.15$  & $70.00$  & $86.92$  \\ \hline

        & \multicolumn{3}{c}{0s}  \\\cline{2-4}
        LLaVA-7B & $10.77$ & $76.92$ & $77.69$ \\
        LLaVA-34B & $13.85$ & $72.31$  & $80.00$ \\
        GPT-4o & $55.38$ & $87.69$ & $99.23$ \\ 
        Gemini-1.5 Pro & $52.31$ & $\mathbf{91.54}$ & $\mathbf{100.00}$ \\
        
        & \multicolumn{3}{c}{0s-CoT} \\\cline{2-4}
        
        LLaVA-7B & $13.85$ & $70.77$ & $84.62$  \\
        LLaVA-34B & $15.38$ & $68.46$ & $83.85$  \\ 
        GPT-4o & $52.31$ & $90.00$ & $\mathbf{100.00}$  \\ 
        Gemini-1.5 Pro & $48.46$ & $\mathbf{91.54}$ & $98.46$  \\ \hline
    \end{tabular}

    \label{tab: overall}
    \end{small}
\end{table}

\section{Analysis and Discussions}

The classification accuracies achieved by various models are presented in Table \ref{tab: overall}. It is important to highlight that in the main experiments, we carefully select machine learning models with embeddings generated by DINOv2, evaluate deep learning architectures (in
\hyperref[Deep Learning Models Comparison]{\textit{Deep Learning Models Comparison}}), and choose the models that demonstrate the best performance on the validation set within their families. For species-level predictions using LLMs, we employ the Name-Listed setting, which results in higher accuracy (further discussed in \hyperref[LLMs' Species Level Settings]{\textit{LLMs' Species Level Settings}}). The best results among the models are highlighted in bold.

\subsection{Main Results Analysis}

SVM performs better than KNN and fine-tuned neural networks across all three levels. Its accuracy is $64.62\%$, $90.00\%$, and $100.00\%$ at species-level, medium-level, and group-level classifications. CLIP achieves an accuracy of $36.15\%$, $70.00\%$, and $86.92\%$ in three levels, respectively.

% However, on average, DL models tend to outperform ML models across different levels. Nonetheless, these ML and DL models provide a solid reference for baseline performance when evaluating LLMs and LLM-MAS, offering a comprehensive benchmark for further comparisons.

% As for LLMs, we use CLIP, LLaVA, GPT-4o, and Gemini-1.5 Pro, have the capability to process both visual and textual information, making them suitable for zero-shot and few-shot classification tasks.

For the LLaVA models, we use the 7B and 34B versions with 4-bit quantization. The model temperature is set to 0 to improve the determinism of the output. Both models exhibit relatively poor performance in the zero-shot setting, achieving accuracies of $10.77\%$ and $13.85\%$, respectively, at the species level. In the zero-shot CoT classification, both models show accuracy improvements at the species and group levels but experience a decline at the medium level. Although the 34B model contains more parameters than the 7B model, no significant performance differences are observed between them. Overall, the LLaVA models consistently rank the lowest in classification accuracy compared to all other models, primarily due to insufficient training at the species level.
 
On the LLM side, surprisingly, we find that at the species level, both GPT-4o and Gemini-1.5 Pro experience a decrease in performance when using CoT, with their accuracy falling short of SVM, VGG, and ResNet, all of which achieve over $60.00\%$. At the medium level, Gemini achieves an accuracy of $91.54\%$ either with or without CoT, which is higher than GPT's performance in the same setting --- $87.69\%$ without CoT and $90.00\%$ with CoT. Therefore, at this level, Gemini demonstrates superior classification capability compared to all other models, and GPT also achieves accuracy equivalent to the best-performing traditional solution (SVM) by using the CoT method. At the group level, both GPT and Gemini outperform all previous approaches except SVM, regardless of whether CoT is used. 

% At the species level, there are a couple of aspects worth noting. First, GPT has an accuracy of $55.38\%$ without the presence of CoT and $52.31\%$ if it is used; similarly, Gemini's performance is $52.31\%$ versus $48.46\%$ with and without CoT. That is, for both models, the accuracy is higher in the zero-shot only setting; the CoT method weakens models' performance at this level.

In conclusion, SVM, VGG, and ResNet achieve higher accuracy scores at the species level than LLMs. This is partly because SVM undergoes training while VGG and ResNet undergo fine-tuning before deployment, allowing them to learn directly from the training set and optimize their performance. Additionally, LLMs' predictions often show a bias toward more common species. This bias stems from the knowledge encoded in the LLMs' structures and weights, which can be distorted by the frequency of species in the pre-training data. Since LLMs are typically pre-trained on vast, diverse multimodal datasets, the prevalence of certain species significantly influences the models' knowledge. Consequently, in zero-shot settings where LLMs rely heavily on their pre-trained knowledge, this bias can skew predictions, favoring species that appear more frequently in the training data.

\begin{table}[th]
\caption{Validation Accuracy of Deep Learning Models}
 \label{tab:dl}
  \renewcommand{\arraystretch}{1.3}
  \tabcolsep 12pt
 \centering
 \begin{tabular}{c c c c} 
 \toprule
  & Species & Medium & Group \\ 
 \midrule 
 \ ResNet34  & $60.76$ & $85.38$ & $97.69$ \\
 \ ResNet50  & $\mathbf{64.62}$ & $88.46$ & $\mathbf{99.23}$ \\
 \ ResNet101 & $60.00$ & $86.92$ & $94.62$ \\
 \ ResNet152 & $63.08$ & $\mathbf{89.23}$ & $98.46$ \\
 \midrule 
 \ VGG11       & $52.30$ & $84.62$ & $95.38$ \\
 \ VGG11 w/ BN & $\mathbf{53.07}$ & $83.85$ & $\mathbf{96.92}$ \\
 \ VGG13       & $52.30$ & $\mathbf{88.46}$ & $96.15$ \\
 \ VGG13 w/ BN & $53.07$ & $84.62$ & $96.92$ \\
 \bottomrule
 \end{tabular}
\end{table}

\subsection{Deep Learning Models Comparison}

\label{Deep Learning Models Comparison}

To identify the best-performing deep learning models, we compare the performance of various variants within the VGG and ResNet families on the validation set. The results are presented in Table \ref{tab:dl}.

\textbf{ResNet:} Among the variants, ResNet50 achieves the highest validation accuracy at the species level ($64.62\%$) and the group level ($99.23\%$), while ResNet152 excels at the medium level with an accuracy of $89.23\%$. However, there is a noticeable drop in species-level performance for the other two ResNet variants, indicating the challenges in distinguishing closely related species.
 
\textbf{VGG:} The VGG architectures exhibit varied performance across our classification tasks. VGG11 with batch normalization (BN) achieves the highest accuracy at the species level ($53.07\%$) and the group level ($96.92\%$). In contrast, VGG13 without batch normalization delivers the best accuracy for the medium level ($88.46\%$), significantly higher than the other variants. This suggests that the impact of batch normalization may vary depending on the specific architecture and task, while it substantially improves performance on this task with moderate complexity.

The results reveal that ResNet and VGG architectures each have distinct strengths depending on the classification level. After comprehensively evaluating the performance of these neural networks on the validation set across all levels, we finally select VGG-11 with BN and ResNet-50 as our baseline models for the main experiments.

% \subsection{Deep Learning Models Comparison}

% \label{Deep Learning Models Comparison}

% Table \ref{tab:dl} shows the validation accuracy of ResNet and VGG families at all levels. The numbers in bold represent the top-performing model architectures, which are subsequently evaluated on the test set.

\subsection{LLMs' Species Level Settings}

\label{LLMs' Species Level Settings}

In Table \ref{tab:Open-Ended}, we list the accuracy of four LLMs using Open-Ended and Name-Listed prompts.

It can be observed that accuracy in the name-listed setting is consistently higher than in the open-ended setting across all models. When the model is provided with possible labels, it no longer needs to generate the species name from scratch or consider a vast range of potential species. Instead, it selects the most likely label from a predefined set, which reduces the cognitive load and leads to more accurate predictions. By focusing on known labels, the model can more effectively compare the input image, resulting in a better performance than when it must explore all possible outputs.

Based on this finding, the main experiments exclusively consider the name-listed setting when using LLMs for predictions.

\begin{table}[th]
 \caption{Open-Ended vs. Name-Listed Settings}
 % \centering
 \renewcommand{\arraystretch}{1.3}
  \tabcolsep 5pt
 \label{tab:Open-Ended}
 \centering
 \begin{tabular}{c c c c c} 
 \toprule
  & LLaVA-7B & LLaVA-34B & GPT-4o & Gemini-1.5 Pro\\ 
 \midrule 
 & \multicolumn{4}{c}{0s}  \\\cline{2-5}
 \ Open-Ended & $3.08$ & $2.31$ & $33.08$ & $43.08$ \\
 \ Name-Listed & $\mathbf{10.77}$ & $\mathbf{13.85}$ & $\mathbf{55.38}$ & $\mathbf{52.31}$ \\
 \midrule 
 & \multicolumn{4}{c}{0s-CoT}  \\\cline{2-5}
 \ Open-Ended & $6.15$ & $5.38$ & $33.85$ & $37.69$ \\
 \ Name-Listed & $\mathbf{13.85}$ & $\mathbf{15.38}$ & $\mathbf{52.31}$ & $\mathbf{48.46}$ \\
 \bottomrule
 \end{tabular}
\end{table}

\subsection{LLM-based MAS Evaluation}
Given the potential strengths of LLM-based MAS, we construct a MAS as illustrated in Figure \ref{fig:multi-agent-system}. The system leverages the outcomes from three foundational LLMs: GPT-4o, Gemini-1.5 Pro, and LLaVA-34B. The predictions from these three agents are fed into the Critic Agent to produce the final results.

The outcome is shown in Table \ref{tab:mas}. At the species level, we observe that the MAS consistently boosts accuracy. Notably, under the zero-shot CoT setting, MAS reaches an accuracy of $56.15\%$ when GPT is used as the ensemble model, outperforming all other LLM settings and narrowing the gap between best-performing LLM-based approaches and traditional counterparts ($60.77\%$ obtained by VGG). At the medium level, Gemini with MAS achieves $92.31\%$ accuracy, regardless of whether CoT is used, which surpasses all non-MAS LLMs and traditional methods, with SVM achieving $90.00\%$. Additionally, there is a slight increase in accuracy at the group level when Gemini is used with MAS under the CoT setting. The results clearly demonstrate that the MAS approach enhances the LLMs' capability across different levels of marine mammal classification.

\begin{table}[th]
\caption{Performance Comparison of MAS}
 \label{tab:mas}
  \renewcommand{\arraystretch}{1.3}
  \tabcolsep 11pt
 \centering
 \begin{tabular}{l l l l} 
 \toprule
  & Species & Medium & Group \\ 
 \midrule 
 % W/o MAS & & & \\
 & \multicolumn{3}{c}{W/o MAS}  \\\cline{2-4}
 % \midrule 
 \ GPT-4o (0s) & $55.38$ & $87.69$ & $99.23$ \\
 \ Gemini-1.5 Pro(0s) & $52.31$ & $91.54$ & $\mathbf{100.00}$ \\
 \ GPT-4o (0s-CoT) & $52.31$ & $90.00$ & $\mathbf{100.00}$ \\
 \ Gemini-1.5 Pro (0s-CoT) & $48.46$ & $91.54$ & $98.46$ \\
 \midrule 
 % W/ MAS & & & \\
 & \multicolumn{3}{c}{W/ MAS}  \\\cline{2-4}
 % \midrule 
 \ GPT-4o(0s) & $55.38$ & $86.92$ & $99.23$ \\
 \ Gemini-1.5 Pro(0s) & $54.62$ & $\mathbf{92.31}$ & $\mathbf{100.00}$ \\
 \ GPT-4o (0s-CoT) & $\mathbf{56.15}$ & $90.00$ & $\mathbf{100.00}$ \\
 \ Gemini-1.5 Pro (0s-CoT) & $50.77$ & $\mathbf{92.31}$ & $99.23$ \\
 \bottomrule
 \end{tabular}
\end{table}

\section{Conclusion}

This paper introduces a specialized marine mammal dataset and conducts a comprehensive evaluation of advanced AI techniques for marine mammal identification and classification. The outcomes show that traditional models such as KNN and SVM, coupled with pre-trained DINOv2 embeddings, excel at fine-grained species-level classification, while LLMs such as GPT-4o and Gemini-1.5 Pro perform well at the medium and group levels. It is important to note that neither traditional models nor LLMs are trained on our dataset. Moreover, the LLM-based MAS further boosts the accuracy compared to a single LLM.

Our results highlight the importance of specialized datasets in advancing AI-driven conservation research. The dataset introduced in this study not only fills a critical gap in marine mammal research but also lays a solid foundation for future developments in AI and conservation efforts. As LLMs continue to evolve, their applications in environmental sciences are expected to enhance our understanding and conservation of marine biodiversity. Future work could explore improving LLMs for species-level tasks, developing more sophisticated multi-agent solutions, and expanding multimodal datasets to further enhance AI capabilities.

\bibliographystyle{IEEEtran}
\bibliography{main}

\begin{thebibliography}{10}
\providecommand{\url}[1]{#1}
\csname url@samestyle\endcsname
\providecommand{\newblock}{\relax}
\providecommand{\bibinfo}[2]{#2}
\providecommand{\BIBentrySTDinterwordspacing}{\spaceskip=0pt\relax}
\providecommand{\BIBentryALTinterwordstretchfactor}{4}
\providecommand{\BIBentryALTinterwordspacing}{\spaceskip=\fontdimen2\font plus
\BIBentryALTinterwordstretchfactor\fontdimen3\font minus \fontdimen4\font\relax}
\providecommand{\BIBforeignlanguage}[2]{{%
\expandafter\ifx\csname l@#1\endcsname\relax
\typeout{** WARNING: IEEEtran.bst: No hyphenation pattern has been}%
\typeout{** loaded for the language `#1'. Using the pattern for}%
\typeout{** the default language instead.}%
\else
\language=\csname l@#1\endcsname
\fi
#2}}
\providecommand{\BIBdecl}{\relax}
\BIBdecl

\bibitem{bowen1997role}
W.~D. Bowen, ``Role of marine mammals in aquatic ecosystems,'' \emph{Marine Ecology Progress Series}, vol. 158, pp. 267--274, 1997.

\bibitem{moore2008marine}
S.~E. Moore, ``Marine mammals as ecosystem sentinels,'' \emph{Journal of Mammalogy}, vol.~89, no.~3, pp. 534--540, 2008.

\bibitem{nelms2021marine}
S.~E. Nelms, J.~Alfaro-Shigueto, J.~P. Arnould, I.~C. Avila, S.~B. Nash, E.~Campbell, M.~I. Carter, T.~Collins, R.~J. Currey, C.~Domit \emph{et~al.}, ``Marine mammal conservation: over the horizon,'' \emph{Endangered Species Research}, vol.~44, pp. 291--325, 2021.

\bibitem{achiam2023gpt}
J.~Achiam, S.~Adler, S.~Agarwal, L.~Ahmad, I.~Akkaya, F.~L. Aleman, D.~Almeida, J.~Altenschmidt, S.~Altman, S.~Anadkat \emph{et~al.}, ``Gpt-4 technical report,'' \emph{arXiv preprint arXiv:2303.08774}, 2023.

\bibitem{gpt4o2024}
OpenAI, ``Gpt-4o: Omni-capable multimodal model,'' \url{https://openai.com/index/hello-gpt-4o/}, 2024.

\bibitem{team2023gemini}
G.~Team, R.~Anil, S.~Borgeaud, Y.~Wu, J.-B. Alayrac, J.~Yu, R.~Soricut, J.~Schalkwyk, A.~M. Dai, A.~Hauth \emph{et~al.}, ``Gemini: a family of highly capable multimodal models,'' \emph{arXiv preprint arXiv:2312.11805}, 2023.

\bibitem{reid2024gemini}
M.~Reid, N.~Savinov, D.~Teplyashin, D.~Lepikhin, T.~Lillicrap, J.-b. Alayrac, R.~Soricut, A.~Lazaridou, O.~Firat, J.~Schrittwieser \emph{et~al.}, ``Gemini 1.5: Unlocking multimodal understanding across millions of tokens of context,'' \emph{arXiv preprint arXiv:2403.05530}, 2024.

\bibitem{touvron2023llama}
H.~Touvron, T.~Lavril, G.~Izacard, X.~Martinet, M.-A. Lachaux, T.~Lacroix, B.~Rozi{\`e}re, N.~Goyal, E.~Hambro, F.~Azhar \emph{et~al.}, ``Llama: Open and efficient foundation language models,'' \emph{arXiv preprint arXiv:2302.13971}, 2023.

\bibitem{touvron2023llama2}
H.~Touvron, L.~Martin, K.~Stone, P.~Albert, A.~Almahairi, Y.~Babaei, N.~Bashlykov, S.~Batra, P.~Bhargava, S.~Bhosale \emph{et~al.}, ``Llama 2: Open foundation and fine-tuned chat models,'' \emph{arXiv preprint arXiv:2307.09288}, 2023.

\bibitem{LLaVA}
Microsoft, ``Llava,'' \url{https://www.microsoft.com/en-us/research/project/llava-large-language-and-vision-assistant/}, 2024.

\bibitem{DALLE3}
OpenAI, ``Dall·e 3,'' \url{https://openai.com/index/dall-e-3/}, 2024.

\bibitem{li2023blip}
J.~Li, D.~Li, S.~Savarese, and S.~Hoi, ``Blip-2: Bootstrapping language-image pre-training with frozen image encoders and large language models,'' in \emph{International conference on machine learning}.\hskip 1em plus 0.5em minus 0.4em\relax PMLR, 2023, pp. 19\,730--19\,742.

\bibitem{CLIP}
OpenAI, ``Clip: Connecting text and images,'' \url{https://openai.com/index/clip/}, 2024.

\bibitem{ALIGN}
G.~Research, ``Align: Scaling up visual and vision-language representation learning with noisy text supervision,'' \url{https://research.google/blog/align-scaling-up-visual-and-vision-language-representation-learning-with-noisy-text-supervision/}, 2024.

\bibitem{gemini1.5pro2024}
G.~Deepmind, ``Gemini pro, our best model for general performance across a wide range of tasks,'' \url{https://deepmind.google/technologies/gemini/pro/}, 2024.

\bibitem{singh2024evaluating}
S.~Singh, G.~Pavlakos, and D.~Stamoulis, ``Evaluating zero-shot gpt-4v performance on 3d visual question answering benchmarks,'' \emph{arXiv preprint arXiv:2405.18831}, 2024.

\bibitem{wang2024llmgeo}
Z.~Wang, D.~Xu, R.~M.~S. Khan, Y.~Lin, Z.~Fan, and X.~Zhu, ``Llmgeo: Benchmarking large language models on image geolocation in-the-wild,'' \emph{arXiv preprint arXiv:2405.20363}, 2024.

\bibitem{pedersen2019detection}
M.~Pedersen, J.~Bruslund~Haurum, R.~Gade, and T.~B. Moeslund, ``Detection of marine animals in a new underwater dataset with varying visibility,'' in \emph{Proceedings of the IEEE/CVF Conference on Computer Vision and Pattern Recognition Workshops}, 2019, pp. 18--26.

\bibitem{wang2023deep}
N.~Wang, T.~Chen, S.~Liu, R.~Wang, H.~R. Karimi, and Y.~Lin, ``Deep learning-based visual detection of marine organisms: A survey,'' \emph{Neurocomputing}, vol. 532, pp. 1--32, 2023.

\bibitem{yang2020cv_aquaculture}
L.~Yang, Y.~Liu, H.~Yu, X.~Fang, L.~Song, D.~Li, and Y.~Chen, ``Computer vision models in intelligent aquaculture with emphasis on fish detection and behavior analysis: A review,'' \emph{Archives of Computational Methods in Engineering}, vol.~27, no.~4, pp. 1107--1132, 2020.

\bibitem{zhang2023sensors}
Y.~Zhang and W.~Gao, ``Research challenges, recent advances, and popular datasets in deep learning-based underwater marine object detection: A review,'' \emph{Sensors}, vol.~23, no.~4, p. 1990, 2023.

\bibitem{ibraheam2021performance}
M.~Ibraheam, K.~F. Li, F.~Gebali, and L.~E. Sielecki, ``A performance comparison and enhancement of animal species detection in images with various r-cnn models,'' \emph{AI}, vol.~2, no.~4, pp. 552--577, 2021.

\bibitem{silva2017cattle}
C.~Silva, D.~Welfer, F.~P. Gioda, and C.~Dornelles, ``Cattle brand recognition using convolutional neural network and support vector machines,'' \emph{IEEE Latin America Transactions}, vol.~15, no.~2, pp. 310--316, 2017.

\bibitem{jian2024underwater}
M.~Jian, N.~Yang, C.~Tao, H.~Zhi, and H.~Luo, ``Underwater object detection and datasets: a survey,'' \emph{Intelligent Marine Technology and Systems}, vol.~2, no.~1, p.~9, 2024.

\bibitem{hong2023metagpt}
S.~Hong, X.~Zheng, J.~Chen, Y.~Cheng, J.~Wang, C.~Zhang, Z.~Wang, S.~K.~S. Yau, Z.~Lin, L.~Zhou \emph{et~al.}, ``Metagpt: Meta programming for multi-agent collaborative framework,'' \emph{arXiv preprint arXiv:2308.00352}, 2023.

\bibitem{wang2023avalon}
S.~Wang, C.~Liu, Z.~Zheng, S.~Qi, S.~Chen, Q.~Yang, A.~Zhao, C.~Wang, S.~Song, and G.~Huang, ``Avalon's game of thoughts: Battle against deception through recursive contemplation,'' \emph{arXiv preprint arXiv:2310.01320}, 2023.

\bibitem{park2023generative}
J.~S. Park, J.~O'Brien, C.~J. Cai, M.~R. Morris, P.~Liang, and M.~S. Bernstein, ``Generative agents: Interactive simulacra of human behavior,'' in \emph{Proceedings of the 36th annual acm symposium on user interface software and technology}, 2023, pp. 1--22.

\bibitem{zeng2024autodefense}
Y.~Zeng, Y.~Wu, X.~Zhang, H.~Wang, and Q.~Wu, ``Autodefense: Multi-agent llm defense against jailbreak attacks,'' \emph{arXiv preprint arXiv:2403.04783}, 2024.

\bibitem{sreedhar2024simulating}
K.~Sreedhar and L.~Chilton, ``Simulating human strategic behavior: Comparing single and multi-agent llms,'' \emph{arXiv preprint arXiv:2402.08189}, 2024.

\bibitem{guo2024embodied}
X.~Guo, K.~Huang, J.~Liu, W.~Fan, N.~V{\'e}lez, Q.~Wu, H.~Wang, T.~L. Griffiths, and M.~Wang, ``Embodied llm agents learn to cooperate in organized teams,'' \emph{arXiv preprint arXiv:2403.12482}, 2024.

\bibitem{guo2024large}
T.~Guo, X.~Chen, Y.~Wang, R.~Chang, S.~Pei, N.~V. Chawla, O.~Wiest, and X.~Zhang, ``Large language model based multi-agents: A survey of progress and challenges,'' \emph{arXiv preprint arXiv:2402.01680}, 2024.

\bibitem{simonyan2014very}
K.~Simonyan, ``Very deep convolutional networks for large-scale image recognition,'' \emph{arXiv preprint arXiv:1409.1556}, 2014.

\bibitem{he2016deep}
K.~He, X.~Zhang, S.~Ren, and J.~Sun, ``Deep residual learning for image recognition,'' in \emph{Proceedings of the IEEE conference on computer vision and pattern recognition}, 2016, pp. 770--778.

\bibitem{LLaVA2023visual}
\BIBentryALTinterwordspacing
H.~Liu, C.~Li, Q.~Wu, and Y.~J. Lee, ``Visual instruction tuning,'' arXiv preprint arXiv:2304.08485, 2023, accessed: 2024-08-27. [Online]. Available: \url{https://arxiv.org/abs/2304.08485}
\BIBentrySTDinterwordspacing

\bibitem{oquab2023dinov2}
M.~Oquab \emph{et~al.}, ``Dinov2: Learning robust visual features without supervision,'' \emph{arXiv preprint arXiv:2304.07193}, 2023.

\bibitem{paszke2019pytorch}
A.~Paszke, S.~Gross, F.~Massa, A.~Lerer, J.~Bradbury, G.~Chanan, T.~Killeen, Z.~Lin, N.~Gimelshein, L.~Antiga \emph{et~al.}, ``Pytorch: An imperative style, high-performance deep learning library,'' \emph{Advances in neural information processing systems}, vol.~32, 2019.

\end{thebibliography}

\end{document}